\documentclass[runningheads]{llncs}

\usepackage{eccv}

\definecolor{dg}{rgb}{0,0.694,0.298}
\definecolor{purple}{rgb}{0.4,0.176,0.569}
\definecolor{royalblue}{RGB}{65,105,225}
\usepackage{pifont}
\usepackage{eccvabbrv}

\usepackage{graphicx}
\usepackage{booktabs}

\usepackage[accsupp]{axessibility}  

\usepackage[pagebackref,breaklinks,colorlinks,citecolor=eccvblue]{hyperref}

\usepackage{orcidlink}
\newcommand{\figref}[1]{Fig.~\ref{#1}}

\newcommand{\secref}[1]{Sec.~\ref{#1}}

\usepackage{graphicx}
\usepackage{amsmath}
\usepackage{amssymb}
\usepackage{booktabs}
\usepackage{multirow}
\usepackage{adjustbox}
\usepackage{sidecap}
\begin{document}

\title{Leveraging Adaptive Implicit Representation Mapping for Ultra High-Resolution Image Segmentation} 


\author{
Ziyu Zhao
\and
Xiaoguang Li
\and
Pingping Cai
\and
Canyu Zhang
\and
Song Wang}

\authorrunning{Ziyu Zhao, Xiaoguang Li, Pingping Cai, Canyu Zhang, Song Wang}

\institute{University of South Carolina, USA \\
\email{\{ziyuz,xl22,pcai,canyu\}@email.sc.edu, songwang@cec.sc.edu}}

\maketitle

\begin{abstract}
  Implicit representation mapping (IRM) can translate image features to any continuous resolution, showcasing its potent capability for ultra-high-resolution image segmentation refinement. Current IRM-based methods for refining ultra-high-resolution image segmentation often rely on CNN-based encoders to extract image features and apply a Shared Implicit Representation Mapping Function (SIRMF) to convert pixel-wise features into segmented results. Hence, these methods exhibit two crucial limitations. Firstly, the CNN-based encoder may not effectively capture long-distance information, resulting in a lack of global semantic information in the pixel-wise features. Secondly, SIRMF is shared across all samples, which limits its ability to generalize and handle diverse inputs.
  %
%
  To address these limitations, we propose a novel approach
  that leverages the newly proposed \textbf{A}daptive \textbf{I}mplicit \textbf{R}epresentation \textbf{M}apping (\textbf{AIRM}) for ultra-high-resolution Image Segmentation. 
  Specifically, the proposed method comprises two components: (1) the \textbf{Affinity Empowered Encoder (AEE)}, a robust feature extractor that leverages the benefits of the transformer architecture and semantic affinity to model long-distance features effectively, and (2) the \textbf{Adaptive Implicit Representation Mapping Function (AIRMF)}, which adaptively translates pixel-wise features without neglecting the global semantic information, allowing for flexible and precise feature translation.
  We evaluated our method on the commonly used ultra-high-resolution segmentation refinement datasets, \ie, BIG and PASCAL VOC 2012. The extensive experiments demonstrate that our method outperforms competitors by a large margin. The code is provided in supplementary material.
\end{abstract}

\begin{figure}[t]
\centering
  \includegraphics[width=0.6\linewidth, height=0.4\linewidth]{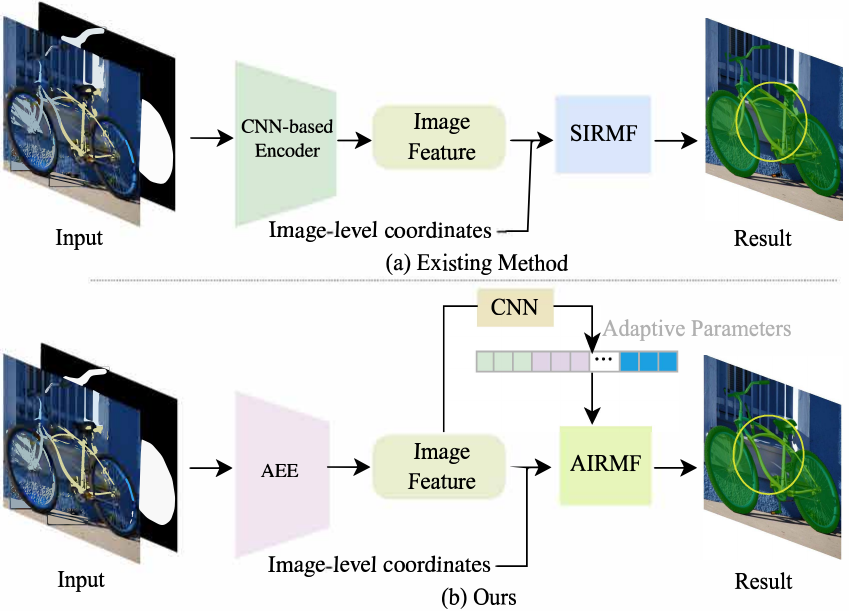} 
              \captionsetup{font={small}}
  \caption{The pipelines for the ultra-high-resolution image segmentation refinement, which take the coarse image and segmentation mask as input. (a) The existing IRM-based methods utilize a Shared Implicit Representation Mapping Function (SIRMF) to transform pixel-wise latent codes into segmentation results. (b) In contrast, our Adaptive Implicit Representation Mapping Function (AIRMF) maps the pixel-wise latent codes adaptively, without neglecting global semantic information.
}
  \vspace{-15pt}
  \label{fig:fig1}
\end{figure}

\section{Introduction}
\label{sec:intro}

Image segmentation, the process of partitioning pixels of an input image into distinct regions, has shown notable progress. While prior methods excel at the low-resolution segmentation task \cite{long2015fully,krizhevsky2012imagenet,simonyan2014very,lin2017feature, chen2017deeplab,chen2016attention}, addressing the task of ultra-high-resolution segmentation still poses a persistent challenge due to substantial computational demands.
%

To address the ultra-high-resolution image segmentation problem, a commonly employed strategy involves refining the segmentation outcomes produced by low-resolution models. Following this strategy, \cite{cheng2020cascadepsp} proposes a refinement network with a cascade scheme, where the intermediate refinement outcomes are iteratively upsampled to higher resolutions.
Despite achieving surprising results, this progressive process is quite time-consuming.
To improve the computational efficiency and reduce the inference time, CRM \cite{shen2022high} draws inspiration from LIIF \cite{chen2021learning} and uses Implicit Representation Mapping (IRM) to achieve the refinement of ultra-high-resolution image segmentation.
While exhibiting promising performance, existing IRM-based methods, \eg, CRM and LIIF, typically rely on a CNN-based encoder to extract image features and a Shared Implicit Representation Mapping Function (SIRMF) to translate the pixel-wise features, 
overlooking two critical limitations: (1) the CNN-based encoder may struggle to effectively capture long-distance information, leading to a lack of global semantic information in pixel-wise features. (2) SIRMF is shared across all samples, limiting its generalization capability to handle diverse inputs.
As a result, these two limitations significantly impact the performance of current IRM-based methods (see the yellow circle in \figref{fig:fig1}).

To address these limitations, our work begins by conducting comprehensive experiments to examine the impact of receptive fields on IRM. We observe that larger receptive fields significantly enhance the performance of IRM. Additionally, we analyze the weaknesses of the existing SIRMF and find that it lacks the generalization capability to handle diverse inputs. Based on these observations, we introduce a novel refinement approach that \textit{Leveraging Adaptive Implicit Representation Mapping (AIRM) for Ultra High-Resolution Image Segmentation}.
The proposed network comprises two components: the Affinity Empowered Encoder (AEE) and the Adaptive Implicit Representation Mapping Function (AIRMF). The AEE, responsible for feature extraction, leverages the advantages of transformer architecture and semantic affinity to model long-distance features effectively. Unlike using shared mapping parameters, AIRMF's parameters are predicted by a separate network based on the entire feature extracted by AEE. This enables AIRMF to adaptively translate pixel-wise features without neglecting global semantic information, allowing for flexible and precise feature translation.
As a result, the proposed method can extract pixel-wise features with a significantly larger receptive field and translate them into the segmentation result in an adaptive manner, collaborating with global semantic information.
Our main contribution is the following.
\begin{itemize}

    \item We conduct comprehensive experiments to explore the impact of receptive fields on IRM and analyze the limitations of the existing SIRMF. The experimental results demonstrate two key findings: (1) Larger receptive fields significantly enhance the performance of IRM, and (2) the existing SIRMF lacks the generalization capability to handle diverse inputs and tends to overlook the global semantic information.

    \item We introduce a novel approach named \textit{Leveraging Adaptive Implicit Representation Mapping for Ultra-High-Resolution Image Segmentation}, which comprises two components: the Affinity Empowered Encoder (AEE) and the Adaptive Implicit Representation Mapping Function (AIRMF). The proposed method can extract pixel-wise features with a significantly larger receptive field and translate them into the segmentation result in an adaptive manner, collaborating with global semantic information.

    \item The extensive experiment results on multiple ultra-high-resolution segmentation refinement datasets demonstrate that our method outperforms competitors by a large margin.
    
\end{itemize}

\section{Related Work}

\subsection{Refining Semantic Segmentation}

The progress in semantic segmentation, which aims to assign distinct categories or classes to individual pixels within an image, has been significantly propelled by the achievements of deep neural networks \cite{long2015fully,krizhevsky2012imagenet,simonyan2014very,lin2017feature, chen2017deeplab,chen2016attention, farabet2012learning, he2004multiscale,mostajabi2015feedforward,shotton2009textonboost,zhang2018context}. 
However, these methods are primarily focused on tackling the issue of low-resolution image segmentation but fall short of producing high-quality segmentation results. 
Thus,
the task of refining segmentation is introduced to improve the quality of segmentation outcomes for high-resolution (1K$\sim$2K) (\textbf{HR}) or even ultra high-resolution  (4K$\sim$6K) (\textbf{UHR}) images. Unlike low-resolution images, \textbf{HR} and \textbf{UHR} require a deeper understanding of semantic details. 
Conventional techniques rely on graphical models like CRF \cite{chen2014semantic,chen2017deeplab,
lin2016efficient, krahenbuhl2011efficient, liu2015semantic} or region growing \cite{dias2019semantic} techniques to achieve HR segmentation. But they mainly focus on low-level color boundaries \cite{zheng2015conditional}, limiting their segmentation performance. 
In contrast, Multi-scale Cascade Networks \cite{chen2019collaborative, cheng2020cascadepsp, he2019bi, qi2018sequential, sun2013deep, zhao2018icnet} employ a recursive approach to progressively integrate multi-scale information. Nevertheless, such a recursive approach inadvertently introduces more computational burdens. Therefore, a Continuous Refinement Network \cite{shen2022high} is proposed to improve the efficiency by continually aligning the feature map with the refinement target. However, it tends to overlook the significance of the limited receptive field of the Convolution Operation in the CNN backbone and also lacks the generalization capability to effectively handle specific categories. To address these concerns, we employ a Transformer-based adaptive implicit representation function to mitigate these limitations and improve segmentation accuracy.

\subsection{Transformer in Segmentation}

%
Vision Transformers (ViT) \cite{dosovitskiy2020image}, as demonstrated by their remarkable success, is adapted for various computer vision tasks such as semantic segmentation \cite{liu2021swin, zheng2021rethinking} and object segmentation \cite{zhu2020deformable, dai2021dynamic}. 
However, applying plain Vision Transformer (ViT) models directly to segmentation tasks is challenging due to their absence of segmentation-specific heads. To solve this problem, SETR \cite{zheng2021rethinking} reframes semantic segmentation as a sequence-to-sequence prediction task. \cite{liu2021swin} proposes a versatile hierarchical Vision Transformer backbone, known as Swin Transformer (Hierarchical Vision Transformer using Shifted Windows), using shifted windows to efficiently capture long-range dependencies while maintaining scalability. Segmenter \cite{strudel2021segmenter} employs pre-trained image classification models fine-tuned on moderate-sized segmentation datasets for semantic segmentation tasks. 
Instead of ViT, \cite{ru2022learning} utilizes Transformers to generate more integral initial pseudo labels for the end-to-end weakly-supervised semantic segmentation task and proposes an Affinity from Attention (AFA) module to learn semantic affinity from the multi-head self-attention (MHSA) in Transformers.
Similar to \cite{ru2022learning}, we propose to integrate the Transformer architecture into our \textbf{UHR} image refinement segmentation task and leverage the learned semantic affinity from multi-head self-attention to enhance the precision of coarse masks.

\subsection{Hypernetworks}

Hypernetworks or meta-models, as referenced in \cite{article, littwin2019deep}, involve designing models specifically tailored for generating parameters for other models. This parameterization approach not only enhances the model's expressiveness, as highlighted in \cite{galanti2020comparing,galanti2020modularity}, but it also supports data compression through weight sharing across a meta-model \cite{article}. 
Unlike conventional deep neural networks, where the weights remain fixed during inference, hypernetworks possess the ability to generate adaptive weights or parameters for target networks \cite{li2023learning, li2022misf, guo2021jpgnet, Li_2023_ICCV, zhang2023superinpaint}. We leverage this technique and propose an Adaptive Implicit Representation Mapping Function. Essentially, an additional Convolution Network is responsible for learning the parameters of the implicit function in the segmentation network,
enhancing its adaptability to a variety of high-resolution image segmentation tasks. 


\begin{figure}[b]
\vspace{-15pt}
\centering
\includegraphics[width=0.8\linewidth]{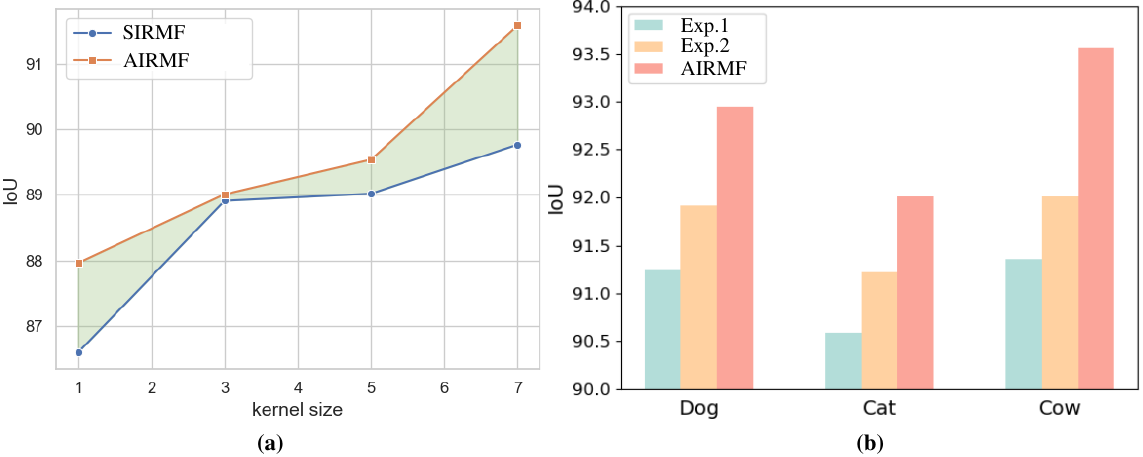}
\captionsetup{font={small}}
\caption{
\vspace{0pt}
(a) The impact of receptive fields on IRM. The blue points and red points denote the SIRMF and AIRMF respectively (b) Analyze the limitation of the existing SIRMF. The blue bar denotes one SIRMF trained on three categories, the yellow bar denotes the three SIRMF trained on each category respectively, and the red bar denotes the AIRFM. 
}
\vspace{0pt}
\label{fig:dis}
\end{figure}
%
\section{Discussion and Motivation}
\label{dis_moti}
%

In this section, we conduct several comprehensive experiments to explore the impact of receptive fields on IRM 
(see \secref{receptive_fields}) and analyze the limitation of the existing SIRMF (see \secref{limitation_ip}). 
We observe that (1) larger receptive fields significantly enhance the performance of IRM, and (2) the existing SIRMF lacks the generalization capability to handle diverse inputs and tends to overlook global semantic information. Building on these observations, we introduce a novel approach named \textit{Leveraging Adaptive Implicit Representation Mapping for Ultra-High-Resolution Image Segmentation}. This approach utilizes transformer architecture and semantic affinity to effectively model long-distance features, enabling adaptive translation of pixel-wise features while preserving global semantic information.
%
\subsection{The Impact of Receptive Fields on IRM}
\label{receptive_fields}
{\bf Image encoder.} To investigate the impact of receptive fields on IRM, we construct a straightforward image encoder, \ie $\phi$, comprising only several convolutional layers, followed by ReLU and batch normalization operations. We then utilize the encoder $\phi$ to extract features $\mathbf{F} \in \mathbb{R}^{h \times w \times D}$ from the input image $\mathbf{I}$ and coarse segmentation mask $\mathbf{M}$,
%
\begin{align}
   \mathbf{F} = \phi(\mathbf{I}, \mathbf{M}).
\end{align}
%

{\bf SIRMF.} Following \cite{chen2021learning}, the SIRMF is implemented by an MLP network, denotes as $\textit{f}_\theta$, and the refined segmentation result $\Tilde{\mathbf{M}}_{q}$ at location ${q}$, can be represented as
%
\begin{align}
    \label{liff}
   \Tilde{\mathbf{M}}_{q} = \sum_{p \in \mathcal{N}_q} \frac{S_p}{\sum{S_p}} \cdot \textit{f}_\theta(z_p, p - q),
\end{align}
%
where $\textit{f}$ is the mapping function, $\theta$ is the trainable parameter of $\textit{f}$,  $\mathcal{N}_q$ are neighbor locations around $q$, and $z_p$ is the pixel-wise feature vector related to location $p$. The aggregation weights, \ie area value of $S_p$, are calculated based on the relative coordinate offset between $q$ and $p$.
%

We train the image encoder $\phi$ with various kernel sizes and the SIRMF $\textit{f}_{\theta}$ on the high-resolution image segmentation dataset. Subsequently, we evaluate the performance based on the Intersection over Union (IoU) and illustrate the results in \figref{fig:dis} (a) (see the blue line). We can observe a substantial performance improvement in the segmentation results when enlarging the kernel size. This highlights that larger receptive fields significantly enhance the performance of IRM in the context of image segmentation tasks. Therefore, in this study, we propose using a transformer-based architecture enhanced with semantic affinity as the image encoder, moving away from the CNN-based encoders commonly used in previous research, \eg, CRM \cite{shen2022high} and LIIF \cite{chen2021learning}.

\subsection{Limitation of The existing SIRMF}
\label{limitation_ip}
To further analyze the limitations of the current SIRMF, we randomly select three categories of images from the segmentation dataset. Subsequently, we conduct the following experiments: 

{\bf Exp.1} We utilize the pre-trained image encoder $\phi$ from \secref{receptive_fields} to extract image features. We train a SIRMF on the entire selected image set and then evaluate the performance of the SIRMF on each category separately (see \figref{fig:dis} (b) blue bar). 

{\bf Exp.2} We use the same encoder as in Exp.1 to extract image features. But we train three SIRMF, each dedicated to one of the selected categories, and then evaluate the performance of each SIRMF on its corresponding category (see \figref{fig:dis} (b) yellow bar). 

From \figref{fig:dis} (b) we can observe that training three SIRMF separately for each category yields better performance compared to training one SIRMF across all three categories. This observation indicates that the existing SIRMF lacks the generalization capability to effectively handle specific categories and tends to overlook global semantic information.

\section{Methodology}
\begin{figure*}[t]
	\begin{center}
	    \includegraphics[width=1\linewidth]{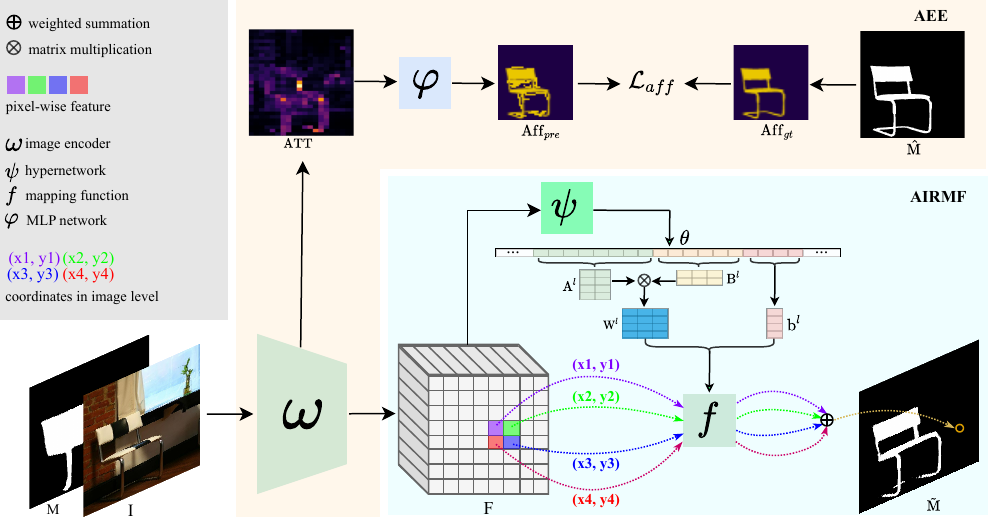} 
            \captionsetup{font={small}}
            \vspace{-20pt}
		\caption{The proposed architecture of \textbf{AIRM}. We employ an affinity-empowered encoder (AEE) to extract the image features. Then we utilize the adaptive implicit representation mapping function (AIRMF) to adaptively translate pixel-wise features without neglecting global semantic information.
  } 
		\label{main}
		\vspace{-20pt}
            \label{pipeline}
	\end{center}

\end{figure*}

%
Building upon the observations in \secref{dis_moti}, in this section, we introduce a novel approach that leverages adaptive implicit representation mapping for ultra-high-resolution image segmentation to effectively address the previous limitations, as shown in \figref{pipeline}. Specifically, the proposed method \textbf{AIRM} comprises two components: the Affinity Empowered Encoder (\textbf{AEE}) (see \secref{encoder}) and the Adaptive Implicit Representation Mapping Function (\textbf{AIRMF}) (see \secref{mapping}). The AEE is responsible for effectively modeling long-distance features by taking advantage of the transformer architecture and semantic affinity. The AIRMF is responsible for adaptively translating pixel-wise features without neglecting the global semantic information, allowing for flexible and precise feature translation.

\subsection{Affinity Empowered Encoder}
\label{encoder}

As discussed in \secref{dis_moti}, larger receptive fields significantly enhance the performance of IRM. Therefore, instead of employing a CNN-based image encoder like CRM \cite{shen2022high} and LIIF \cite{chen2021learning}, we propose an affinity-empowered image encoder that takes advantage of the transformer architecture and semantic affinity \cite{ru2022learning}.
Specifically, given the RGB image $\mathbf{I} \in \mathbb{R}^{H \times W \times 3}$ and the corresponding coarse segmentation mask $\mathbf{M} \in \mathbb{R}^{H \times W \times 1}$, we employ the transformer-based encoder $\omega$ to extract features $\mathbf{F}$, which can be represented as follows:
\begin{equation}
    \mathbf{F}, \mathbf{ATT} = \omega(\mathbf{I}, \mathbf{M}),
\end{equation}
where $\mathbf{ATT} \in \mathbb{R}^{hw \times hw \times c}$ denotes the attention map produced by $\omega$ and $hw$ represents the flattened spatial dimensions and $c$ is the number of attention heads. Note that the coarse segmentation mask is obtained by applying an existing segmentation method such as \cite{long2015fully, chen2018encoder, lin2017refinenet}.
%


To further empower the capability of the transformer-based encoder, we follow \cite{ru2022learning} to learn the affinity from the attention map $\mathbf{ATT}$ to intentionally impose constraints on the encoder $\omega$ during training. 
The self-attention mechanism functions as a directed graphical model, where the attention $\mathbf{ATT}$ is typically coarse (see \figref{pipeline}). The affinity matrix is expected to be symmetric since nodes with identical semantics should be equal.
Therefore, following \cite{ru2022learning}, we obtain the predicted affinity $\mathbf{Aff}_{pre}\in \mathbb{R}^{hw \times hw}$ (see \figref{pipeline}) by feeding the sum of $\mathbf{ATT}$ and its transpose into an MLP network, \ie, $\varphi$, which can be represented as: 
\begin{equation}
    \mathbf{Aff}_{pre} = \varphi(\mathbf{ATT} + \mathbf{ATT}^\top).
\end{equation}
We expect $\mathbf{Aff}_{pre}$ to be close to the ground truth affinity $\mathbf{Aff}_{gt} \in \mathbb{R}^{hw \times hw}$ obtained from the ground truth segmentation mask $\hat{\mathbf{M}}$. Specifically, we resize $\hat{\mathbf{M}}$ to $hw \times hw$, and the affinity between two coordinates $(x_i, y_i)$ and $(x_j, y_j)$ is assigned as 1 if their classes are identical, and 0 otherwise.

\subsection{Adaptive Implicit Representation Mapping Function (AIRMF)}
\label{mapping}

In \secref{limitation_ip}, we identified the limitation in the existing SIRMF \eqref{liff} — it lacks the capacity to effectively generalize across diverse inputs (see \figref{fig:dis} (b)). To tackle this issue, we introduce SIRMF which can be represented as
%
\begin{align}
    \label{our_liff}
   \Tilde{\mathbf{M}}_{q} = \sum_{p \in \mathcal{N}_q} \frac{S_p}{\sum{S_p}} \cdot \textit{f}_{\psi(\mathbf{F})}(z_p, p - q),
\end{align}
%
where $\psi(\cdot)$ denotes a hypernetwork that is used to predict the adaptive mapping parameters.
Instead of directly optimizing the shared parameter $\theta$ in \eqref{liff}, we employ $\psi$ to predict it adaptively by inputting the feature $\mathbf{F}$. Intuitively, the predicted mapping parameter should capture global semantic information from the input feature $\mathbf{F}$. 
Consequently, AIRMF can leverage both local information from pixel-wise features and global information from the entire image feature. This design allows AIRMF to yield precise segmentation results (see the red points in \figref{fig:dis} (a) and the red bar in \figref{fig:dis} (b)).

Assume we aim to generate the weight ${\textit{\textbf{W}}^\textit{l}} \in \mathbb{R}^{n_{\rm in} \times n_{\rm out}}$ and bias ${\textit{\textbf{b}}^\textit{l}} \in \mathbb{R}^{n_{\rm out}}$ of the \textit{l}-th linear layer of $f(\cdot)$. To optimize computational efficiency, we do not directly predict the parameter. Instead, we adopt a strategy similar to \cite{skorokhodov2021adversarial}, predicting two intermediary rectangular  matrices ${\textit{\textbf{A}}^\textit{l}} \in \mathbb{R}^{n_{\rm out}^l \times r}$  and ${\textit{\textbf{B}}^\textit{l}} \in \mathbb{R}^{{r \times n_{\rm in}^l}}$  through $\psi(\cdot)$, where r = 20 for all the layers of $\textit{f}{(\cdot)}$. The linear modulating projection weight ${\textit{\textbf{W}}}^l$ is then modulated by multiplying these matrices as follow, 
\begin{equation}
    {\textit{\textbf{W}}}^l = {\textit{\textbf{A}}^\textit{l}} \times {\textit{\textbf{B}}^\textit{l}}.
\end{equation}
Bias parameter ${\textit{\textbf{b}}^\textit{l}}$ is directly produced due to its small volume. We streamline the AIRMF in \figref{pipeline}.



\subsection{Implementation Details}

{\bf Network architectures.} We employ the Mix Transformer (MiT) introduced in Segformer \cite{xie2021segformer} as the backbone of image encoder $\omega$. MiT is a more suitable backbone for image segmentation tasks compared to the vanilla ViT \cite{dosovitskiy2020image}. In brief, MiT incorporates overlapped patch embedding to maintain local consistency, spatial-reductive self-attention to accelerate computation, and employs FFN with convolutions to safely replace position embedding. The network of $\varphi$ to predict the semantic affinity is implemented by an MLP layer. 
The deep network $\psi$ employed to generate the flattened parameters of the mapping function $\textit{f}$ is constructed by several \textit{Conv+ReLU+BatchNorm} layers, followed by a linear layer.

{\bf Loss.}
During training, we use the combined loss functions to optimize overall performance. To improve AIRMF, we utilize the cross-entropy loss $\mathcal{L}_{ce}$, L1 loss $\mathcal{L}_{1}$, L2 loss $\mathcal{L}_{2}$ and gradient loss $\mathcal{L}_{grad}$
\begin{equation}
    \mathcal{L}_{AIRMF} = \lambda_1\mathcal{L}_{1} + \lambda_2\mathcal{L}_{2} + \lambda_3\mathcal{L}_{ce} +\lambda_4\mathcal{L}_{grad},
\end{equation}
where $\lambda_1=0.2$, $\lambda_2=0.2$, $\lambda_3=0.3$ and $\lambda_4=0.4$ balance the contributions in $\mathcal{L}_{AIRMF}$. The ablation study to evaluate each component of $\mathcal{L}_{AIRMF}$ is included in Sec. \ref{abla}.


To calculate the affinity loss $\mathcal{L}_{aff}$, we concentrate on adjacent coordinates during training to leverage their sufficient context while minimizing computational costs. We use a search mask on $\hat{\mathbf{M}}$ with a predefined radius $R$ as a threshold, efficiently limiting the distance between selected coordinate pairs. This approach ensures that only pixel pairs within the same local search mask are considered, excluding the affinity between distant pixel pairs. The affinity loss term is formulated as follows:

\begin{equation}
\begin{aligned}
    \mathcal{L}_{aff} &= - \frac{1}{\lvert \mathcal{S}^{-} \rvert} \sum_{(i,j) \in \mathcal{S}^{-}} \log{(1-\mathbf{Aff}_{pre}^{i,j})} 
    - \frac{1}{\lvert \mathcal{S}^{+} \rvert} \sum_{(i,j) \in \mathcal{S}^{+}} \log {(\mathbf{Aff}_{pre}^{i,j})},
\end{aligned}
\end{equation}
where $\mathcal{S}^{+}$ and $\mathcal{S}^{-}$ refer to two subsets of positive and negative affinity pixel pairs: 
\begin{equation}
    \mathcal{S}^{+} = \{{(i,j) \mid (i,j) \in \mathcal{P}, \mathbf{Aff}_{pre}^{i,j} = 1}\},
\end{equation}
\begin{equation}
    \mathcal{S}^{-} = \{{(i,j) \mid (i,j) \in \mathcal{P}, \mathbf{Aff}_{pre}^{i,j} = 0}\}.
\end{equation}
Our final loss can be written as:
\begin{equation}
\begin{aligned}
    \mathcal{L}= \frac{1}{2}\mathcal{L}_{AIRMF} + \frac{1}{2}\mathcal{L}_{aff}.
\end{aligned}
\end{equation}

{\bf Training details.} During training, we utilize the Adam optimizer \cite{kingma2017adam} with an initial learning rate $2.25 \times 10^{-5}$ to train our network. The learning rate is decayed to one-tenth at iterations 22,500 and 37,500 in a total of 45,000 iterations. We randomly crop input with a resolution of 224x224 from the original images. During the training, the coarse masks are generated by introducing random perturbations to the ground truth masks with a random IoU threshold ranging from 0.8 to 1.0. The radius $R$ of the local window size when computing the affinity loss is set to 14.

We follow the multi-resolution inference strategy proposed in \cite{shen2022high}, which can be seen as progressively coarse-to-fine operations. Specifically, the inference begins at a very low resolution around the training image resolution and gradually increases the input’s resolution along the continuous ratio axis $Rs \in (0.125,0.25,0.5,1)$. At each refinement stage, the refined mask is continuously concatenated as a coarse mask with input images for the subsequent refinement step.

\section{Experiments}
%
\subsection{Setups}
{\bf Datasets.} To acquire object information and evaluate our models, we adhere to a dataset setup similar to that of CascadePSP \cite{cheng2020cascadepsp}. Our training dataset comprises a combination of several individual datasets, totaling 36,572 RGB images accompanied by their corresponding ground truth masks. These images encompass a wide spectrum of over 1,000 categories, drawing from MSRA-10K \cite{6871397}, DUT-OMRON \cite{6619251}, ECSSD \cite{7182346}, and FSS-1000 \cite{Wei2019FSS1000A1}. For testing, CascadePSP introduced a high-resolution image segmentation dataset, named BIG, ranging from 2K to 6K. Additionally, recognizing that the commonly used original PASCAL VOC 2012 dataset lacks pixel-perfect segmentations and designates regions near object boundaries as 'void', \cite{cheng2020cascadepsp} relabeled 500 objects to create a new dataset suitable for the task of \textbf{UHR} image segmentation.

{\bf Metrics.} We evaluate our model using Intersection over Union (IoU) and mean Boundary Accuracy (mBA) \cite{cheng2020cascadepsp}, which robustly computes the segmentation accuracy across different image sizes within a specified radius from the ground truth boundary. Our model is evaluated without resorting to finetuning in various settings, demonstrating the enhancements it achieves.


\begin{table*}[tb]
    \caption{Comparison of IoU and mBA results on BIG and PASCAL VOC 2012 datasets with varied mask refinement methods, utilizing coarse masks generated by FCN, RefineNet,  DeepLabV3+, and PSPNet.}
    \vspace{-8pt}
 \renewcommand{\arraystretch}{0.93}
  \centering
  \setlength{\tabcolsep}{12pt}
	\begin{adjustbox}{width=\linewidth,center}
        \begin{tabular}{cccccc}
            \toprule
            IoU/mBA(\%)&Coarse Mask&  SegFix \cite{yuan2020segfix}    & CascadePSP \cite{cheng2020cascadepsp} & CRM \cite{shen2022high}& AIRM(Ours)\\
            \midrule
            \multicolumn{6}{c}{\centering \textbf{BIG}}\\
            \midrule
            FCN-8s \cite{long2015fully}    &72.39/53.63 & 72.69/55.21  & 77.87/67.04& 79.62/69.47& \color{red}81.59/\color{red}70.98\\
            RefineNet \cite{chen2018encoder} &90.20/62.03 & 90.73/65.95  & 92.79/74.77& 92.89/75.50& \color{red}93.59/\color{red}77.26\\
            DeepLabV3+ \cite{lin2017refinenet}&89.42/60.25 & 89.95/64.34  & 92.23/74.59& 91.84/\color{red}74.96& \color{red}{93.99}/\color{black}74.77\\
            PSPNet \cite{zhao2017pyramid}   &90.49/59.63 & 91.01/63.25  & 93.93/75.32& 94.18/76.09& \color{red}95.95/\color{red}76.88\\
            \midrule
            \multicolumn{6}{c}{\centering \textbf{PASCAL VOC 2012}}\\
            \midrule
            FCN-8s \cite{long2015fully}    &68.85/54.05 & 70.02/57.63  & 72.70/65.36& 73.34/67.17& \color{red}74.59/\color{red}67.26\\
            RefineNet \cite{chen2018encoder}&87.13/61.68 & 86.71/66.15  & 87.48/71.34& 87.18/71.54& \color{red}88.03/\color{red}71.89\\
            DeepLabV3+ \cite{lin2017refinenet}&87.13/61.18 & 88.03/66.35  & 89.01/72.10& 88.33/\color{red}72.25& \color{red}{91.47}/\color{black}72.01\\
            PSPNet \cite{zhao2017pyramid}    &90.92/60.51 & 91.98/66.03  & 92.86/72.24& 92.52/72.48& \color{red}94.06/\color{red}74.39\\
            \bottomrule
        \end{tabular}
	\end{adjustbox}

        \label{quantitative}
        \vspace{-18pt}
\end{table*}

\subsection{Quantitative Comparison}

 Table \ref{quantitative} provides a comprehensive comparison of class-agnostic IoU and mBA among our \textbf{AIRM}, SegFix \cite{yuan2020segfix}, CascadePSP \cite{cheng2020cascadepsp} and CRM \cite{shen2022high}. All models are trained on low-resolution images and evaluated on their original high-resolution counterparts. Across both datasets, our method consistently outperforms other methods in terms of IoU and mBA. It's worth noting that SegFix's refinement performances are comparatively lower, as it is not specifically designed for the UHR image segmentation task. In the BIG dataset, our method exhibits an average improvement of around 1.7\% for IoU and 1.2\% for mBA based on four fundamental segmentation approaches. Similarly, in the relabeled PASCAL VOC 2012 dataset, our method achieves an average increase of approximately 1.1\% for IoU and 0.6\% for mBA compared to CRM and CascadePSP upon four approaches. These quantitative findings underscore the effectiveness of our model in capturing global semantic features and exploring pixel-wise affinity interrelationships in \textbf{UHR} image segmentation.
 
\begin{figure}[t]
\centering
    {\includegraphics[width=1.0\linewidth]{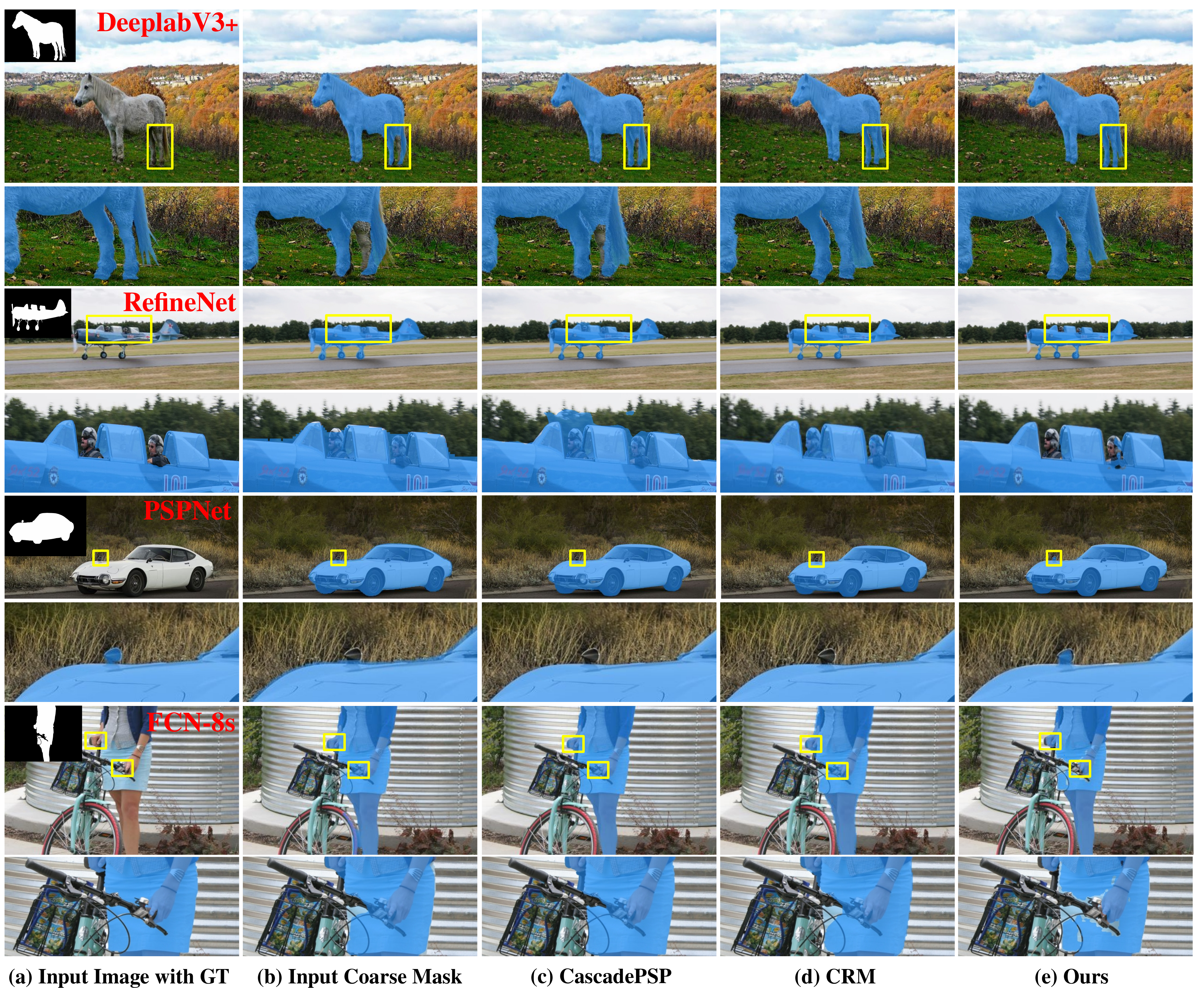}}
    \vspace{-15pt}
    \caption{Qualitative comparison of CascadePSP, CRM and \textbf{AIR} on the coarse masks from DeepLabV3+, RefineNet, PSPNet and FCN-8s. The images are from BIG dataset (2K$\sim$6K). The binary mask in the top-left part of the first column represents the ground truth. Please zoom in for clear details.}
    \label{big}
    \vspace{-12pt}
\end{figure}

\subsection{Qualitative Comparison}

In \figref{big}, we present a qualitative comparison of the performance of CascadePSP, CRM, and our \textbf{AIRM}. Remarkably, even without any prior exposure to high-resolution training images, our method consistently produces refined results of exceptional quality across various scales. When employing four distinct input coarse masks, it becomes evident that our approach excels at reconstructing missing parts in comparison to CascadePSP and CRM. For instance, as depicted in \figref{big}, our method accurately distinguishes the pilot and the passenger seated in the rear of the airplane (the second row) from the segmented object. Additionally, our method precisely delineates the shape of the right rear-view mirror (the third row), achieving a notably high level of precision when compared to the ground truth, which other methods fail to accomplish.
\begin{figure}[!h]
  \centering
\includegraphics[width=0.7\linewidth,height=0.20\textwidth]{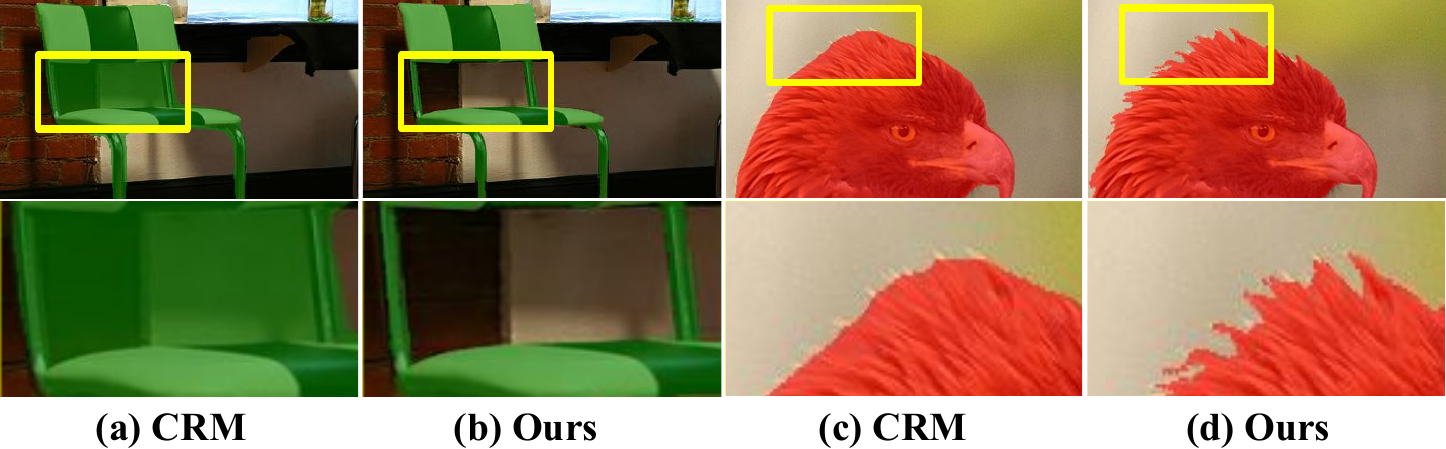}

   \caption{Qualitative output between CRM and our refinement algorithm on PASCAL VOC 2012 dataset(left to right). Coarse masks are from DeepLabV3+.}
   \label{Voc}
   \vspace{-15pt}
\end{figure}

Furthermore, \figref{Voc} underscores the substantial visual enhancements achieved by our refinement technique on the relabeled PASCAL VOC 2012 dataset, which consists of low-resolution images. A direct comparison highlights the generation of finer details in our outputs, particularly along the boundary region, such as the gap in the back of the chair and intricate feather textures. More results in supplementary material further demonstrate the efficacy of our refinement algorithm.

\subsection{Ablation Study}
\label{abla}

\begin{table}
   \vspace{-20pt}
  \caption{ The ablation study of each component of AEE. R50, R101 and Mit-b1 denote ResNet50, ResNet101 \cite{he2016deep}, and Mix Transformer respectively. The results are evaluated on the BIG dataset. Coarse masks are from DeepLabV3+.}
    \vspace{-5pt}
  \centering
  \setlength{\tabcolsep}{12pt}
    \begin{adjustbox}{width=0.6\linewidth,center}
  \begin{tabular}{ccc|c|c|c}
    \toprule
     R50 & R101 & Mit-b1 & Affinity & AIRMF & IoU/mBA \\
    \midrule
    \ding{52}& & &  & \ding{52} &  88.99/67.96 \\
    \linespread{1.5}
    \ding{52}& & & \ding{52} & \ding{52} & 90.14/68.29 \\
    \midrule
    &\ding{52} & &  & \ding{52} &  89.53/68.99 \\    
    \linespread{1.5}
    & \ding{52}& & \ding{52} & \ding{52} & 90.57/69.19 \\    
    \midrule
    & & \ding{52}&  & \ding{52} &  91.25/72.96 \\  
    \linespread{1.5}
    & &\ding{52} & \ding{52} &\ding{52} & \color{red}93.99/74.77 \\      
    \bottomrule
  \end{tabular}
  \end{adjustbox}
  \label{w/o}
\vspace{-15pt}
\end{table}

\noindent
{\bf Affinity Empowered Encoder (AEE).} For simplicity, We follow the method in \cite{ahn2018learning} to acquire the affinity prediction map in CNN-based backbone. The semantic affinity between a pair of feature vectors is defined in terms of their L1 distance. From Table \ref{w/o}, we can find that: \ding{182} Utilizing the transformer-based backbone yields superior segmentation results compared to employing the CNN-based backbone. Specifically, employing Mit-b1 as the image encoder enhances performance by 2.37\%/2.17\% compared to employing R50. \ding{183} Affinity learning proves beneficial for both CNN-based and transformer-based backbones. Performance improves by 2.89\%/1.21\% when using R50 and by 2.59\%/2.14\% when using Mit-b1. \ding{184} Our method, empowered by affinity learning and a transformer-based backbone, achieves the best results.
%
%
%

\begin{figure}[!h]
  \centering
\includegraphics[width=0.8\linewidth,height=0.3\textwidth]{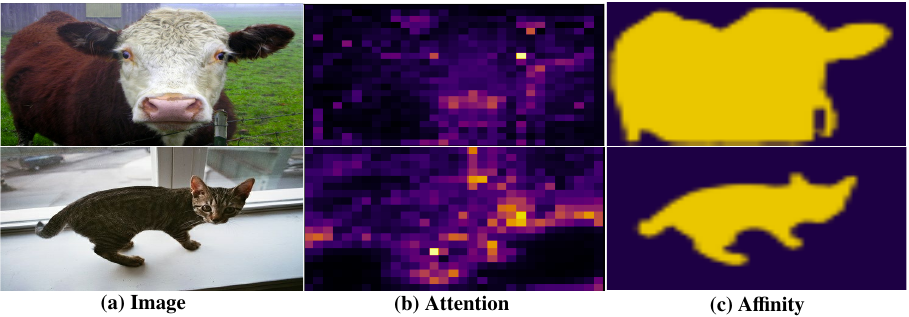}

   \caption{Visualization of attention map and affinity. }
   \label{affinity}
   \vspace{-8pt}
\end{figure}

In addition, we also provide visualized attention maps and learned affinity in \figref{affinity}. From the visualizations, we observe that the attention map (\figref{affinity}(b)) only captures coarse-level semantic information, whereas the learned affinity (\figref{affinity}(c)) provides more reliable semantic information. This outcome strongly validates both our initial motivation and the effectiveness of the AEE module.

\begin{figure}[bt]

\centering
\includegraphics[width=0.45\linewidth,height=0.3\linewidth]{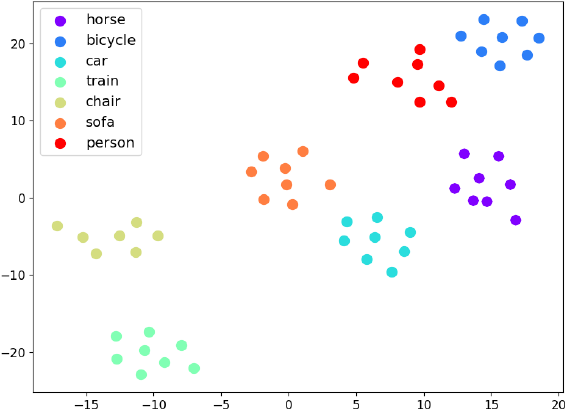}
\vspace{-5pt}
\caption{Visualization of the adaptive parameters.}
\vspace{-20pt}
\label{adaptive}
\end{figure}
\noindent
{\bf Adaptive Implicit Representation Mapping Function (AIRMF).} To demonstrate the effectiveness of AIRMF, we replace it with SIRMF and present the comparison results in Table \ref{sir,air}. We observe that without the proposed AIRMF, the final segmentation results notably decreases across all evaluation metrics, such as the IoU decreases by 1.45 and mBA decreases by 4.00.
%

In addition, we employ the TSNE \cite{van2008visualizing} to reduce the dimension of the predicted adaptive parameters and illustrate that in \figref{adaptive}. We observe that for input images with a similar semantic appearance, the corresponding mapping parameters are closely related. This demonstrates that the predicted adaptive parameters in AIRMF possess the capability to capture global semantic information.

\begin{table}[h]
\vspace{-15pt}
\caption{The ablation study of AIRMF.}
  \vspace{-10pt}
  \centering
  \setlength{\tabcolsep}{12pt}
  \begin{adjustbox}{width=0.5\linewidth,center}
  \begin{tabular}{c|cc|c}

    \toprule
    \linespread{1.0}
    AEE & SIRMF & AIRMF &IoU/mBA   \\
    \midrule
    \linespread{1.5}
     \ding{52} & \ding{52} & & 92.54/70.77  \\
    \linespread{1.5}
     \ding{52} &  &\ding{52} & \color{red}93.99/74.77 \\
    \bottomrule
  \end{tabular}
  \end{adjustbox}
  \label{sir,air}
  \vspace{-20pt}
\end{table}



\begin{table}[h]
\vspace{0pt}
  \caption{Computation cost. FLOPs and time are tested on one image (2560*1706) on average. }
\vspace{-5pt}
\renewcommand{\arraystretch}{0.8}
  \centering
  \setlength{\tabcolsep}{12pt}
      \begin{adjustbox}{width=0.8\linewidth,center}
  \begin{tabular}{cccc|c|ccc}

    \toprule
    R50&Mit-b1 & SIRMF & AIRMF&IoU& FLOPs(G) & Time(s) & Params(M)  \\
    \midrule
    \ding{52}& &\ding{52}& & 88.75 & {1525} & 4.25 & 9.27\\
    \linespread{1.5}
    \ding{52}& &&\ding{52} & 89.08 & 1869 & 4.50 & 9.27\\
    \linespread{1.5}
    &\ding{52}& \ding{52}& & 92.54&1905& 5.14 & 13.21 \\
    \linespread{1.5}
    &\ding{52}&  & \ding{52} & 93.99& 2897& 5.32 &  13.21\\
    \midrule
    \multicolumn{4}{c|}{CasPSP \cite{cheng2020cascadepsp}} & 91.23&26518& 6.20 & 67.62\\
    \multicolumn{4}{c|}{CRM \cite{shen2022high}}           & 91.56&2536& 4.35&9.27\\
    \bottomrule
  \end{tabular}
        \end{adjustbox}
  \label{cost}
  \vspace{-20pt}
\end{table}

\noindent
{\bf Computation Cost.} We conduct experiments to evaluate the time consumption and the model parameters of each component of our method and the competitors, as shown in Table \ref{cost}. We can find that: \ding{182} 
When comparing Transformer and CNN-based backbones, despite an increase in the number of parameters by around 3M (Lines 1 and 3, 2 and 4), our network's exceptional performance remains evident, with the corresponding increase in time being below 1.0\%. \ding{183} In comparing AIRMF and SIRMF, the inference time and number of parameters remain nearly identical, as we employ a simple hypernetwork to achieve adaptive effectiveness; however, the results significantly outperform expectations.\ding{184} When compared to previous methods such as cascade networks (CasPSP) and conventional implicit functions (CRM), our method significantly enhances precise results in \textbf{UHR} image semantic segmentation by around 4.1\% and 4.0\%, respectively, while maintaining computation costs well within an acceptable range.

\noindent
{\bf loss.}
\begin{table}[h]
    \vspace{-15pt}
  \caption{The ablation study on different loss functions. }
    \vspace{-8pt}
\renewcommand{\arraystretch}{0.8}
  \centering
  \setlength{\tabcolsep}{12pt}
       \begin{adjustbox}{width=0.5\linewidth,center}
  \begin{tabular}{ccccc}
    \toprule
    $\mathcal{L}_{1}$ \& $\mathcal{L}_{2}$ & $\mathcal{L}_{ce}$& $\mathcal{L}_{grad}$ & IoU & Time(H)\\
    \midrule
    \ding{52}& \ding{52}& &93.17  & 15.7\\
        \linespread{1.5}
    \ding{52}& & \ding{52}&93.48  & 15.5\\
        \linespread{1.5}
    & \ding{52}& \ding{52}&93.85  & 14.9\\
        \linespread{1.5}
    \ding{52}& \ding{52}& \ding{52}&93.99  & 15.8\\
    \bottomrule
  \end{tabular}
  \end{adjustbox}
  \vspace{-20pt}
  \label{AIR}
\end{table}
We conduct an ablation study on the loss function by individually removing each of its components. As shown in Table \ref{AIR}, the performance decreases with the exclusion of each component. Moreover, different combinations of the loss only slightly affect training time.

\noindent
{\bf Search mask radius in affinity loss.}
In Table \ref{radius}, we also study the segmentation performance on different search mask radius $R$ when generating affinity label for computing affinity loss. Through comparison, we observe that a small value of $R$ may not yield sufficient affinity pairs, whereas a large 
$R$ might compromise the reliability of distant affinity pairs. Thus, we choose $R$ = 14 as our priority.

\begin{table}[h]
        \vspace{-15pt}
      \caption{Ablation study on radius $R$ of the affinity loss. Results are tested on VOC dataset. Coarse masks are from DeepLabV3+.}
         \vspace{-8pt}
	\centering
	\renewcommand\arraystretch{0.8}
 \centering
	\setlength{\tabcolsep}{12pt}
        \begin{adjustbox}{width=0.5\linewidth,center}
	\begin{tabular}{c|c|c|c|c|c}
           \toprule
		    $R$ & 6 & 10 & \color{red}14 & 18 & 20\\
            \midrule
		    IoU& 91.28&91.38& \color{red}91.47&90.99&90.85\\
            \bottomrule
    	\end{tabular}
     \end{adjustbox}
      \label{radius}
        \vspace{-40pt}
\end{table}

\section{Conclusion}

 
Our study extensively examines how receptive fields impact implicit representation mapping and addresses the limitations of current shared mapping functions. We introduce an innovative framework for Ultra High-Resolution (UHR) image semantic segmentation, comprising the Affinity Empowered Encoder (AEE) and the Adaptive Implicit Representation Mapping Function (AIRMF). This methodological approach facilitates the extraction of pixel-wise features with a much larger receptive field, translating it into a segmentation result enriched with global semantic information. Our meticulous experiments demonstrate that our method significantly outperforms competitors.

\textbf{Limitations} While our proposed method outperforms all competitors, the current approach still adheres to the standard configuration of Ultra High-Resolution Image Segmentation testing, which requires a coarse segmentation mask as input, leading to computational burden. Our forthcoming efforts will focus on resolving this intricate issue, emphasizing the robustness of the framework and enhancing its capability to directly handle high-resolution images.
\bibliographystyle{splncs04}
\bibliography{main}
\end{document}